\documentclass{article}

\usepackage[round,semicolon]{natbib}

    \usepackage[final]{neurips_data_2022}

\usepackage[compact]{titlesec}
\usepackage[utf8]{inputenc} %
\usepackage[T1]{fontenc}    %
\usepackage{hyperref}       %
\hypersetup{
    colorlinks=true,
  linkcolor=blue,
  filecolor=magenta,      
  urlcolor=blue,
  citecolor=black,
  pdfinfo={
    Title={How Would The Viewer Feel? Estimating Wellbeing From Video Scenarios},
    Author={Mantas Mazeika, Eric Tang, Andy Zou, Steven Basart, Jun Shern Chan, Dawn Song, David Forsyth, Jacob Steinhardt, Dan Hendrycks},
    Subject={Recommender Systems, Affective Computing, Machine Ethics},
    Keywords={emotions, wellbeing, preferences, alignment, ml safety, human values}
  }
}
\usepackage{url}            %
\usepackage{booktabs}       %
\usepackage{amsfonts}       %
\usepackage{nicefrac}       %
\usepackage{microtype}      %
\usepackage{xcolor}         %

\usepackage{enumitem}

\usepackage{cite}
\usepackage{graphicx}

\usepackage{tikz}
\usepackage{comment}
\usepackage{amsmath,amssymb} %
\usepackage{color}

\usepackage{graphicx}
\usepackage{amsmath}
\usepackage{amssymb}
\usepackage{booktabs}
\usepackage{xcolor}
\usepackage{bbm}
\DeclareMathOperator*{\argmax}{arg\,max}

\DeclareMathOperator*{\argsort}{arg\,sort}
\usepackage{wrapfig}

\usepackage[noabbrev,nameinlink,capitalize]{cleveref}
\crefname{section}{Sec.}{Secs.}
\Crefname{section}{Section}{Sections}
\Crefname{table}{Table}{Tables}
\crefname{table}{Tab.}{Tabs.}

\makeatletter
\newcommand{\printfnsymbol}[1]{%
  \textsuperscript{\@fnsymbol{#1}}%
}
\makeatother

\title{How Would The Viewer Feel?\\Estimating Wellbeing From Video Scenarios}

\author{%
  Mantas Mazeika\thanks{Equal Contribution.} \\
  UIUC
  \And
  Eric Tang\printfnsymbol{1} \\
  UC Berkeley
  \And
  Andy Zou \\
  UC Berkeley
  \And
  Steven Basart \\
  UChicago
  \And
  Jun Shern Chan \\
  UC Berkeley
  \AND
  Dawn Song \\
  UC Berkeley
  \And
  David Forsyth \\
  UIUC
  \And
  Jacob Steinhardt \\
  UC Berkeley
  \And
  Dan Hendrycks \\
  UC Berkeley
}

\begin{document}

\maketitle

\begin{abstract}
In recent years, deep neural networks have demonstrated increasingly strong abilities to recognize objects and activities in videos. However, as video understanding becomes widely used in real-world applications, a key consideration is developing human-centric systems that understand not only the content of the video but also how it would affect the wellbeing and emotional state of viewers. To facilitate research in this setting, we introduce two large-scale datasets with over $60,\!000$ videos manually annotated for emotional response and subjective wellbeing. The Video Cognitive Empathy (VCE) dataset contains annotations for distributions of fine-grained emotional responses, allowing models to gain a detailed understanding of affective states. The Video to Valence (V2V) dataset contains annotations of relative pleasantness between videos, which enables predicting a continuous spectrum of wellbeing. In experiments, we show how video models that are primarily trained to recognize actions and find contours of objects can be repurposed to understand human preferences and the emotional content of videos. Although there is room for improvement, predicting wellbeing and emotional response is on the horizon for state-of-the-art models. We hope our datasets can help foster further advances at the intersection of commonsense video understanding and human preference learning.\looseness=-1
\end{abstract}

\section{Introduction}

Videos are a rich source of data that depict vast quantities of information about humans and the world. As deep learning has progressed, models have begun to reliably exhibit various aspects of video understanding, including action recognition \citep{Kay2017TheKH}, object tracking \citep{Zhao2021GeneratingMF}, segmentation \citep{Huang2019CCNetCA,He2020MaskR}, and more. However, vision models do not exist in a vacuum and will eventually require social perception abilities, so models need to begin understanding how humans interpret and respond to visual inputs. As video models become more widely used in real-world applications, they should be able to reliably predict not only ``what is where'' in a visual input but also predict how it would make a human feel.

The subjective experience of human viewers on video data is broadly valuable to characterize and predict. When humans pursue goals in the world, their actions are often driven by intuitive processes \citep{Kahneman2011ThinkingFA}, a significant part of which is the experience of emotions or affective states \citep{Oatley2006UnderstandingE2}. Emotions can be thought of as evaluations of events in relation to goals \citep{Scherer2001AppraisalPI,Frijda1988TheLO}, and hence are important to study in relation to behavior in diverse settings. However, they are also important to understand in their own right, as they are strong indicators of what people value \citep{hume}. For example, if a situation makes one feel happy, then that is often preferred to a situation that induces feelings of fear. Additionally, emodiversity---the variety of experienced emotions---is an important indicator of the overall health of the human emotional ecosystem, and emodiversity over positive emotions can improve mental wellbeing \citep{quoidbach2014emodiversity}. Thus, understanding the emotional responses and preferences of humans on video data could be a useful avenue toward modeling basic human desires, values, and overall wellbeing.

Video recommender systems already attempt to capture human preferences over videos but for practical reasons often base their recommendations on imperfect proxy metrics \citep{Ridgway1956DysfunctionalCO}. It is hard to directly measure the values of users and how video content affects their wellbeing. Thus, recommender systems often rely on metrics that are easier to obtain, such as engagement and watch time. This simplifies the problem but can result in unintended consequences and safety concerns \citep{hendrycks2021unsolved}. Simplifying metrics loses sight of the experiencer \citep{Scott1999SeeingLA} and can result in situations where engagement is maximized but users are unhappy \citep{Russell2019HumanCA,kross2013facebook,fbupdate,Stray2020AligningAO,Stray2021WhatAY}. For instance, content that evokes feelings of envy or anger can be highly engaging but is nonetheless unhealthy to be constantly exposed to. Thus, systems that recommend videos could substantially improve user experience through content-based inferences about how it would affect the emotional state and wellbeing of viewers.

\begin{figure*}[t]
    \centering
    \includegraphics[width=\textwidth]{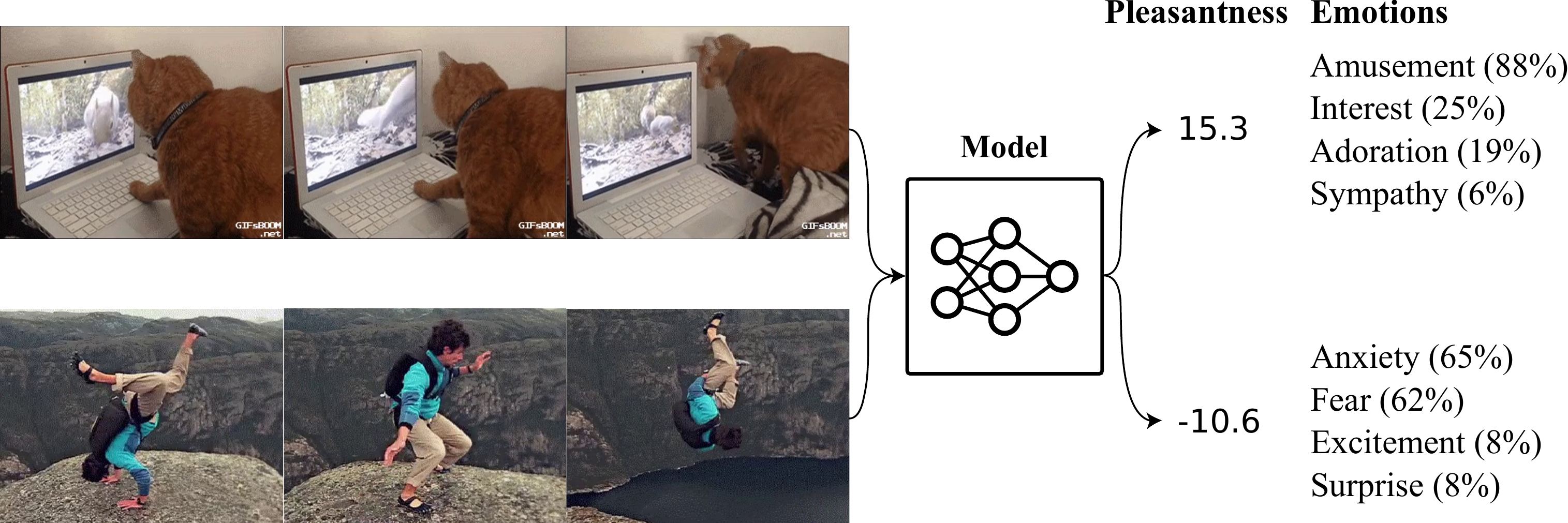}
    \caption{
    We introduce two large-scale datasets for predicting subjective responses to videos, including relative pleasantness between videos and distributions of fine-grained emotional responses. This enables training state-of-the-art vision models to predict continuous, consistent scores for the pleasantness of videos and a rich distribution of likely emotional responses.
    }
    \vspace{-10pt}
    \label{fig:splash}
\end{figure*}

To facilitate research on understanding how viewers feel while watching videos, we introduce two large-scale datasets for predicting emotional state and wellbeing of viewers directly from videos. First, we introduce the Video Cognitive Empathy (VCE) dataset for predicting fine-grained emotional responses to videos. The VCE dataset contains approximately $60,\!000$ videos with human annotations for $27$ emotion categories, ranging from the six basics (joy, sadness, fear, disgust, anger, surprise) \citep{Ekman1992AnAF} to more nuanced emotions such as admiration and awkwardness, altogether covering the spectrum of affective states \citep{Cowen2017SelfreportC2}. As emotional responses can be considered evaluations of events in relation to a person's unique goals, they can vary significantly across human viewers. To capture the diversity of human responses, we collect a distribution---not just a single label---of emotional responses for each video. This enables evaluating models on their ability to inclusively predict the likely range of responses to a video across our large pool of annotators.

To estimate how videos affect the wellbeing of human viewers, we introduce a second dataset, Video to Valence (V2V). The V2V dataset contains approximately $25,\!000$ videos with human-annotated rankings of pleasantness between videos. Pleasantness captures the overall positive or negative affect that viewers feel when watching a video and serves as a measure of wellbeing \citep{sidgwick_1907,LazariRadek2017UtilitarianismAV}. Since our annotations are for pairwise or listwise comparisons across videos, we can train utility-style models to predict continuous wellbeing scores \citep{Hendrycks2021AligningAW}, capturing gradations of wellbeing rather than a binary indicator. For instance, two scary videos may both be unpleasant, but our dataset enables predicting which video is more unpleasant, enabling a deeper understanding of human preferences.

Our datasets come with strong baselines. We train state-of-the-art video Transformers \citep{Vaswani2017AttentionIA} on our tasks and find that these models, which are primarily used for understanding the literal content of videos, can predict the subjective state of viewers with surprising reliability. Although there is room for improvement, models that understand how viewers feel when watching videos are on the horizon and may thus prove useful in numerous applications.
Our datasets and experiment code can be found at {\color{blue} \href{ https://github.com/hendrycks/emodiversity}{github.com/hendrycks/emodiversity}}. We hope our datasets can help foster further research into the important problem of understanding human emotions and wellbeing.

\begin{table*}[t]
\setlength\tabcolsep{4pt}
\begin{center}
\begin{tabular}{l c c c c c}
\hline
Dataset & Annotation Type & Number of Videos \\
\hline
COGNIMUSE \citep{zlatintsi2017cognimuse} & affective labels & $7$ \\
HUMAINE \citep{douglas2007humaine} &  affective labels & $50$ \\
FilmStim \citep{schaefer2010assessing} & affective labels & $70$ \\
DEAP \citep{koelstra2012deap} & affective labels, face video & $120$ \\
VideoEmotion \citep{jiang2014predicting} & discrete emotions & $1,\!101$ \\
LIRIS-ACCEDE \citep{baveye2015liris} & valence, arousal & $160$ \\
EEV \citep{Sun2020EEVDP} & performative expressions & $5,\!153$ \\
\hline
Video Cognitive Empathy (Ours) & fine-grained emotions & $61,\!046$ \\
Video to Valence (Ours) & relative pleasantness & $26,\!670$ \\
\hline
\end{tabular}
\end{center}
\caption{Comparisons between datasets for predicting the subjective states that human viewers would feel while watching videos. We introduce two new datasets with substantially more scenarios than prior work. Our datasets are annotated with subjective self-reports, enabling high-quality evaluations.}
\label{table:dataset}
\vspace{-10pt}
\end{table*}

\section{Related Work}

\noindent\textbf{Video Understanding With DNNs.}\quad
Much work in video understanding has focused on identifying various aspects of the scenarios depicted in videos. These include recognizing human motion and actions \citep{schuldt2004recognizing, kuehne2011hmdb, soomro2012ucf101, wang2014action, karpathy2014large, caba2015activitynet, abu2016youtube, kay2017kinetics, goyal2017something, zhang2019comprehensive}, arbitrary event recognition \citep{monfort2019moments}, spatial localization and tracking \citep{yilmaz2006object, milan2016mot16, kang2016review, vondrick2018tracking}, and video segmentation \citep{pont20172017, xu2018youtube, garcia2018survey}. Some work focuses on recognizing emotions and goals expressed by humans in videos, including facial emotion recognition \citep{lyons1998coding, lucey2010extended, bargal2016emotion, li2020deep} and recognizing unintended actions \citep{epstein2020oops}. Numerous video models have been proposed and benchmarked on tasks for understanding ``what is where'' in videos \citep{gorelick2007actions, tran2015learning, tran2018closer, feichtenhofer2019slowfast, sharir2021image}. However, relatively little work has investigated the context in which videos are often consumed---namely, that humans watch videos and have subjective experiences deriving from said videos. Our work focuses on this important, less explored area of study.

\vspace{10pt}\noindent\textbf{Predicting Subjective Responses.}\quad
Predicting the subjective responses of humans to various stimuli is an important topic of study spanning numerous fields. The International Affective Picture System (IAPS) \citep{lang2007international} and Open Affective Standardized Image Set (OASIS) \citep{kurdi2017introducing} both contain approximately $1,\!000$ images selected to evoke a range of emotional responses. \citet{achlioptas2021artemis} explore affective explanations of paintings as a source of training for deep learning. Eliciting emotions in text is harder, although many works have investigated predicting emotions expressed by writing \citep{strapparava2007semeval, oberlander2018analysis, demszky2020goemotions}. Unlike still images and text, video is better suited to studying subjective responses, as video stimuli can be far more evocative. Numerous datasets have been proposed to study emotional responses to video \citep{zlatintsi2017cognimuse, douglas2007humaine, schaefer2010assessing, koelstra2012deap, jiang2014predicting, baveye2015liris, Sun2020EEVDP}. Notably, \citet{Cowen2017SelfreportC2} collect self-reported emotional states on a bank of $2,\!185$ online videos and find that reported emotional states factor into $27$ distinct emotions, which we use as a framework for building our VCE dataset, which is $30\times$ larger. Comparisons of our datasets to existing work are given in Table \ref{table:dataset}. Our datasets have a much greater scale and diversity of videos than prior work, enabling research on predicting subjective responses with state-of-the-art deep learning models.

\vspace{10pt}\noindent\textbf{Value Learning.}\quad
Building machine learning systems that interact with humans and pursue human values may require understanding aspects of human subjective experience. Many argue that values are derived from subjective experience \citep{hume,sidgwick_1907,LazariRadek2017UtilitarianismAV} and that some of the main components of subjective experience are emotions and valence. Learning representations of values is necessary for creating safe machine learning systems \citep{hendrycks2021unsolved} that operate in an open world. In natural language processing, models are trained to assign wellbeing or pleasantness scores to arbitrary text scenarios \citep{Hendrycks2021AligningAW}. Recent work in machine ethics \citep{Anderson2011MachineE} has translated this knowledge into action by using wellbeing scores to steer agents in diverse environments \citep{Hendrycks2021WhatWJ}. However, this recent line of work so far exclusively considers text inputs rather than raw visual inputs.

\vspace{10pt}\noindent\textbf{Emodiversity.}\quad
A large body of work in psychology seeks to understand and quantify the richness and complexity of human emotional life \citep{barrett2009variety, lindquist2008emotional, carstensen2000emotional}. An important concept in this area is emodiversity, the variety and relative abundance of emotions experienced by an individual, which has been linked with reduced levels of anxiety and depression \citep{quoidbach2014emodiversity}. Although prior work studies emodiversity in self-reports of emotion without stimuli, we hypothesize that the emodiversity of visual stimuli may be an important concept to quantify and understand. Thus, we investigate how our new datasets enable measuring the emodiversity of in-the-wild videos on a large scale.

\begin{figure*}[t]
    \centering
    \includegraphics[width=\textwidth]{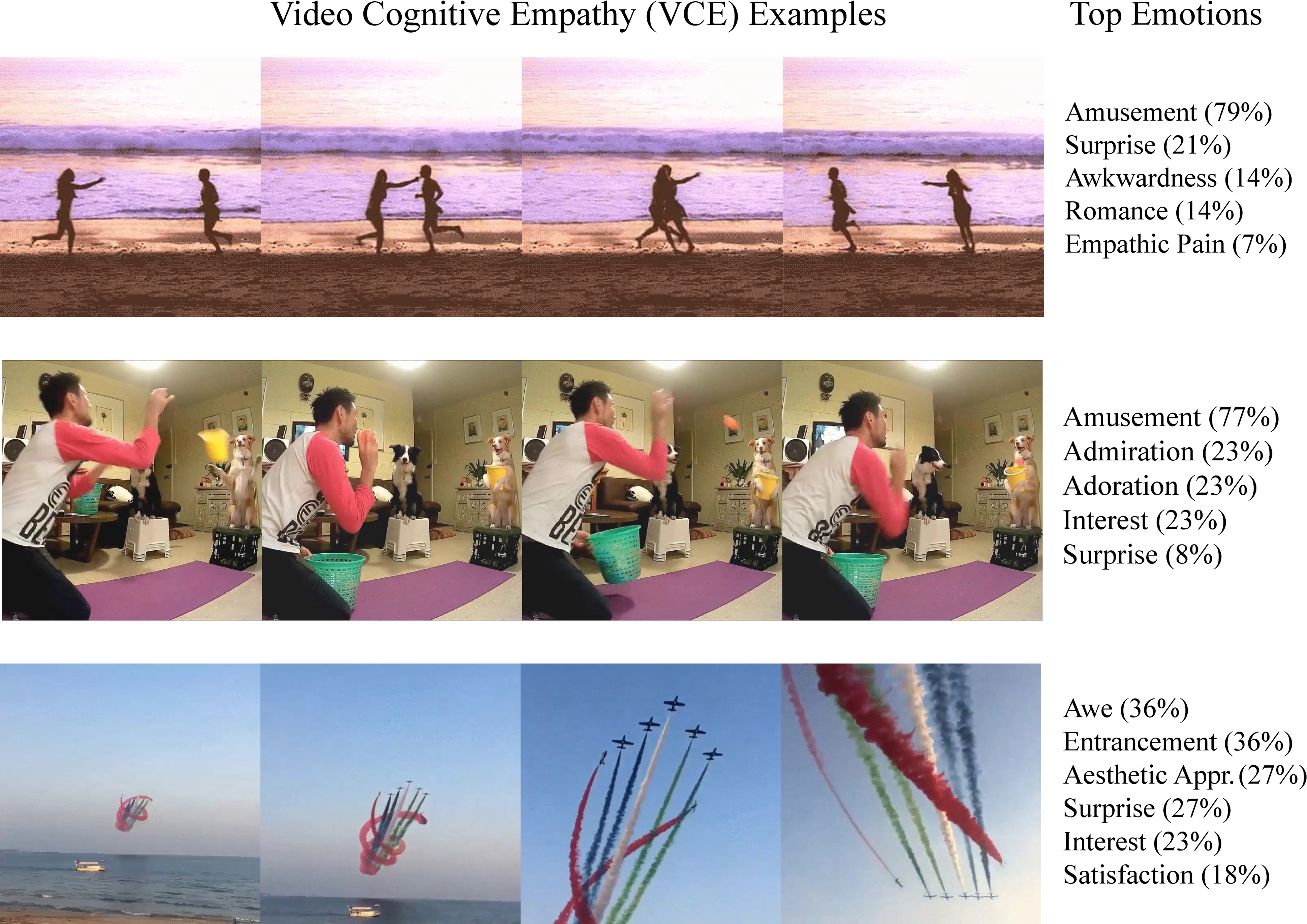}
    \caption{Examples from the Video Cognitive Empathy (VCE) dataset. Each video is annotated with a distribution of emotional responses from forced choice decisions across multiple annotators. We ask whether models can predict emotional responses solely from the semantic content of videos.}
    \vspace{-10pt}
    \label{fig:vce}
\end{figure*}

\section{Video Cognitive Empathy (VCE) Dataset}
When watching videos, humans feel a wide range of emotions based on the semantic content depicted in the video. These emotional responses may depend on the video in complex ways, requiring reasoning about the implications of depicted events as well as a robust understanding of human values. We are interested in whether deep models can exhibit cognitive empathy, the ability to understand how someone else is feeling or would feel in a certain situation. To test whether state-of-the-art video models can predict emotional responses, we introduce the Video Cognitive Empathy (VCE) dataset.

\vspace{10pt}\noindent\textbf{Dataset Description.}
The VCE dataset contains $61,\!046$ videos with annotations for the emotional response of human viewers. The data are split into a training and test set of $50,\!000$ and $11,\!046$ videos, respectively. Each video lasts an average of $14.1$ seconds for a total of $239$ hours of manually annotated data. While movies often evoke emotions with soundtracks and appropriate choices of colors and lighting, we are interested in how emotions depend on the semantic content of videos and less so on how engineered cues can evoke desired emotions. Thus, we remove audio cues that could serve as confounding variables. We also filter out inappropriate videos using an automated nudity detector followed by manual filtering for each video based on video thumbnails. VCE is the first dataset of its size with manual annotations that is suitable for evaluating modern deep video models.

The annotations in VCE are modeled after the analysis performed by \citet{Cowen2017SelfreportC2}. By collecting reported emotional experiences from humans on a set of $2,\!000$ videos, they find that emotional responses exhibit $27$ dimensions associated with reliably distinct situations. These correspond to $27$ descriptive emotional states, such as ``admiration'', ``anger'', and ``amusement''. We adopt this fine-grained categorization of emotions and ask annotators to indicate which emotions they felt the most while watching a video. In Figure 2 of the Supplementary Material, we show the number of annotations per emotion. %

As emotional responses can vary across annotators, we capture the distribution of responses by gathering a large number of annotations per video. For each video in VCE, we gather an average of $13$ annotations (minimum of $12$, maximum of $15$). Rather than only keeping examples with high inter-annotator agreement, which would result in a small dataset, we consider the distribution of responses to be the target for learning. This is justified because while individual emotional responses are variable, the distribution of emotional responses tends to change with the stimuli. For example, scary movies might not scare everyone, but the dominant response is fear. However, responses to certain content such as political videos can vary considerably across populations. Hence, our annotations should not be taken to be representative of all emotional responses and are primarily intended for studying whether deep networks can acquire cognitive empathy.

\begin{figure}
    \centering
    \includegraphics[width=0.8\textwidth]{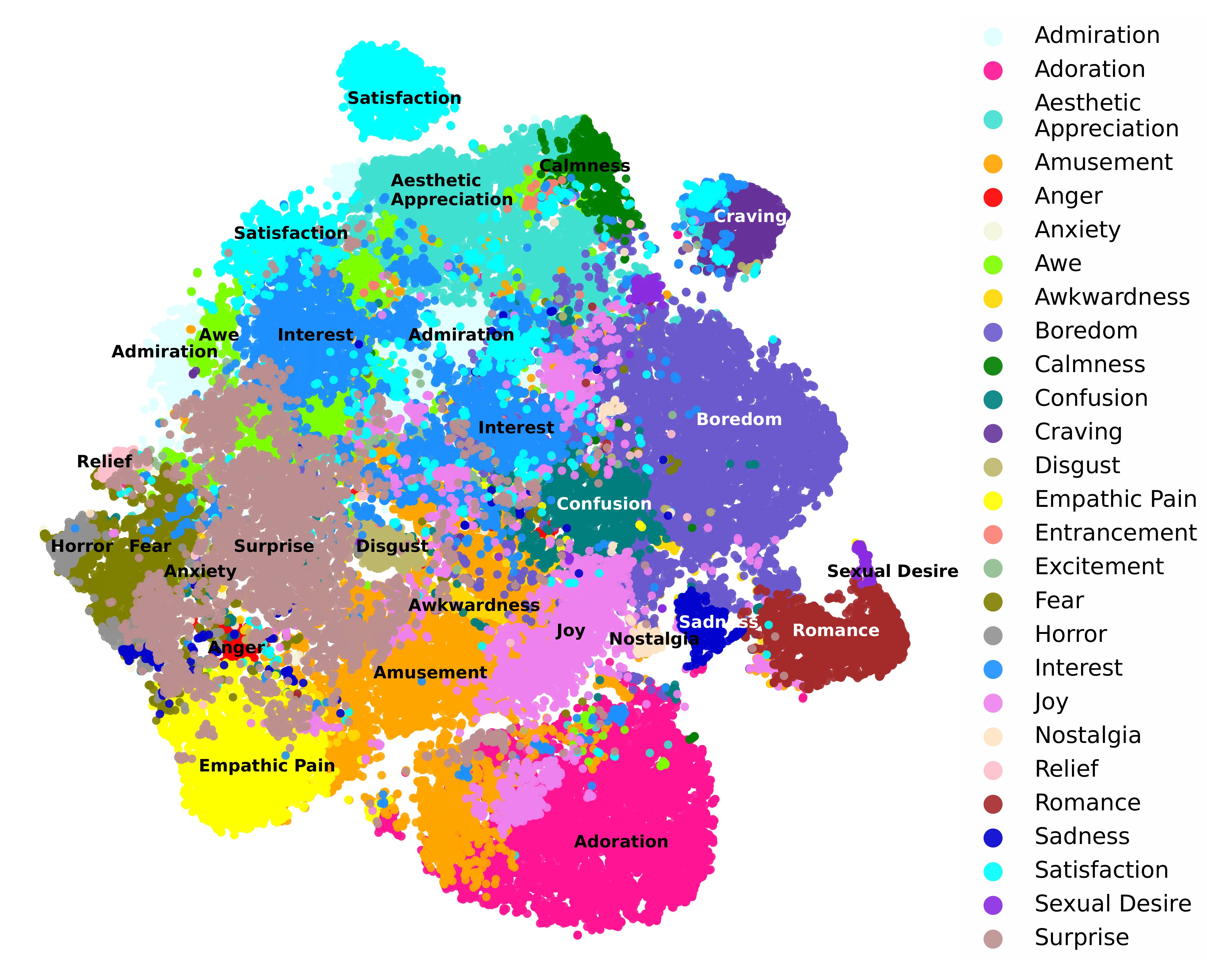}
    \caption{t-SNE plot of all $27$-dimensional annotation vectors in the Video Cognitive Empathy dataset. Points are colored according to the most prevalent evoked emotion. Groups of emotions cluster together in natural ways, allowing for intuitively reasonable traversals through the space of emotions.
   }
   \vspace{-10pt}
    \label{fig:tsne}
\end{figure}

\vspace{10pt}\noindent\textbf{Dataset Construction.}
Annotations for VCE were collected using Amazon Mechanical Turk (MTurk) with IRB approval. For each video, workers were asked to view the video without audio and select from the set of 27 emotions the emotions that the video most strongly evoked. For each selected emotion, workers were asked to rank the intensity of that emotion from 1 to 10. To ensure that labels are high quality, we required that MTurkers pass a qualification test, and provided them with detailed definitions of each of the 27 emotions. We also ensured that workers viewed the entire video, only worked on one task at a time, and asked workers to mark videos that would rely too heavily on audio in order to rate.

Note that the MTurk annotators and individuals depicted in the videos may not form a representative sample of diverse cultural backgrounds. Hence, our annotations should not be taken to represent accurate emotional responses across a broad range of cultures or on an individual level, and we discourage their use in deployment contexts. The VCE and V2V datasets are designed to give a high-level understanding of how well current video models can predict subjective responses to videos. We support work on large-scale data collection that considers differences in emotional responses across cultures and individuals, and we think this is an interesting direction for future research.

\subsection{Metrics}
We evaluate models on VCE using a top-$k$ accuracy metric. Let $(x, y) \in \mathcal{D}$ be a sample video and annotation. The annotation $y$ is a $27$-dimensional vector with non-negative entries that indicates the intensity of responses for each of the $27$ emotion categories. Let $f(x)$ be the predicted output distribution of a model $f$ on video $x$. The top-$k$ accuracy is computed as $\frac{1}{\left| \mathcal{D} \right|} \sum_{(x, y) \in \mathcal{D}} \mathbbm{1}\left[ \argmax f(x) \in \left[\argsort y\right]_{-k:} \right]$, where argsort is in ascending order and the colon notation indicates the last $k$ indices of the resulting array. This measures the fraction of test examples where the maximum predicted emotion is in the top $k$ emotions of the ground-truth distribution. We set $k=3$ for our evaluations.

\subsection{Analysis}
\noindent\textbf{Emotion Clusters.}
\citet{Cowen2017SelfreportC2} find that emotions vary continuously and cluster in reasonable ways. For example, one can smoothly traverse their $27$-dimensional space of emotions by going from calmness to aesthetic appreciation to awe. To investigate whether our responses exhibit this behavior, we perform dimensionality reduction on the $27$-dimensional VCE response distribution using t-SNE. We visualize results in \cref{fig:tsne}. Points are colored according to the maximum emotion in the response distribution. We find that emotions cluster together and that clusters group in natural ways. The groupings exhibit smooth transitions similar to \citet{Cowen2017SelfreportC2}. For example, one can smoothly transition through $\text{calmness} \to \text{aesthetic appreciation} \to \text{awe}$,\quad and\quad $\text{adoration} \to \text{amusement} \to \text{surprise}$. This demonstrates that the distributions of emotional responses contain significant hidden information beyond the top emotion for a given video.

\begin{wrapfigure}{r}{0.54\textwidth}
    \vspace{-15pt}
    \centering
    \includegraphics[width=0.52\textwidth]{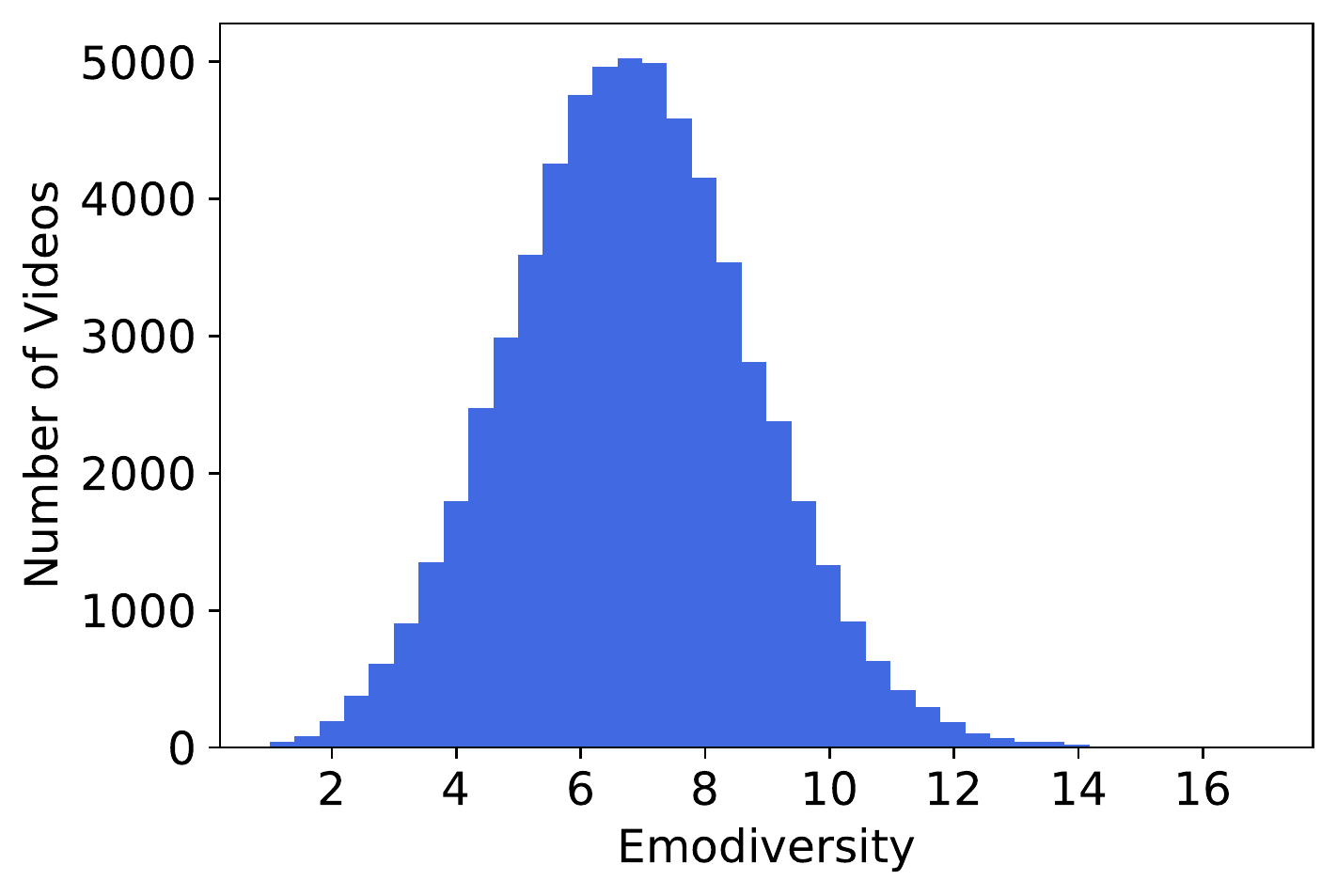}
    \caption{The emodiversity of videos in VCE. The value of the emodiversity can be interpreted as the number of emotions that a video evokes. Most videos evoke a wide range of emotions across viewers.}
    \label{fig:emodiversity}
    \vspace{-10pt}
\end{wrapfigure}

\vspace{10pt}\noindent\textbf{Emodiversity.}
The emotion distribution labels in VCE enable measuring per-video emodiversity. Emodiversity is an indicator of the overall health of the human emotional ecosystem and is positively correlated with mental wellbeing \citep{quoidbach2014emodiversity}. Prior work measures emodiversity using the Shannon entropy \citep{quoidbach2014emodiversity}, but this metric can be hard to interpret \citep{magurran2003measuring}. Thus, we quantify emodiversity using perplexity of the normalized emotion distribution, computed as $\exp \left( -1 \cdot \sum_i y_i \log y_i \right)$ where $y_i$ is the normalized probability assigned to the $i^\text{th}$ emotion in video label $y$. One may also exclude negative emotions like disgust when computing emodiversity, since emodiversity over positive emotions is more relevant to improving wellbeing. This metric has a minimum of $1$ and a maximum of the number of emotion clusters. It can be interpreted as the number of emotions that a video evokes, assuming uniform responses for all evoked emotions. In \cref{fig:emodiversity}, we show the distribution of emodiversity across VCE. This shows that most videos evoke a diverse range of emotions across the population of viewers.

\begin{figure*}[t]
    \centering
    \includegraphics[width=\textwidth]{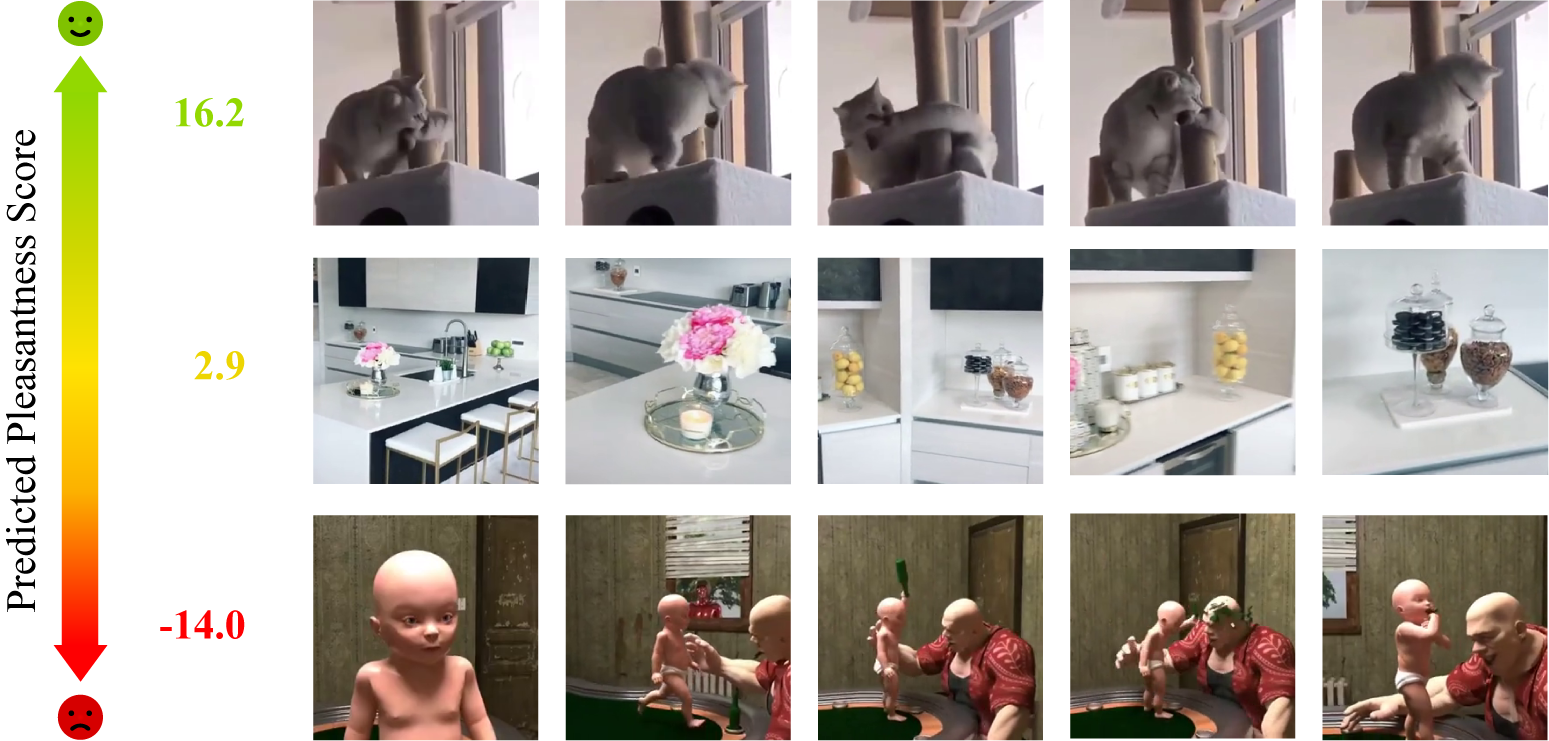}
    \caption{Example predictions on the V2V dataset. We train video models to predict continuous pleasantness scores by enforcing consistency with pleasantness rankings in the V2V training set. This results in intuitively reasonable outputs that capture preferences over the content depicted in videos.
    }
    \label{fig:utility_example}
\end{figure*}

\section{Video to Valence (V2V) Dataset}

\begin{wrapfigure}{r}{0.55\textwidth}
    \vspace{-12pt}
    \centering
    \includegraphics[width=0.53\textwidth]{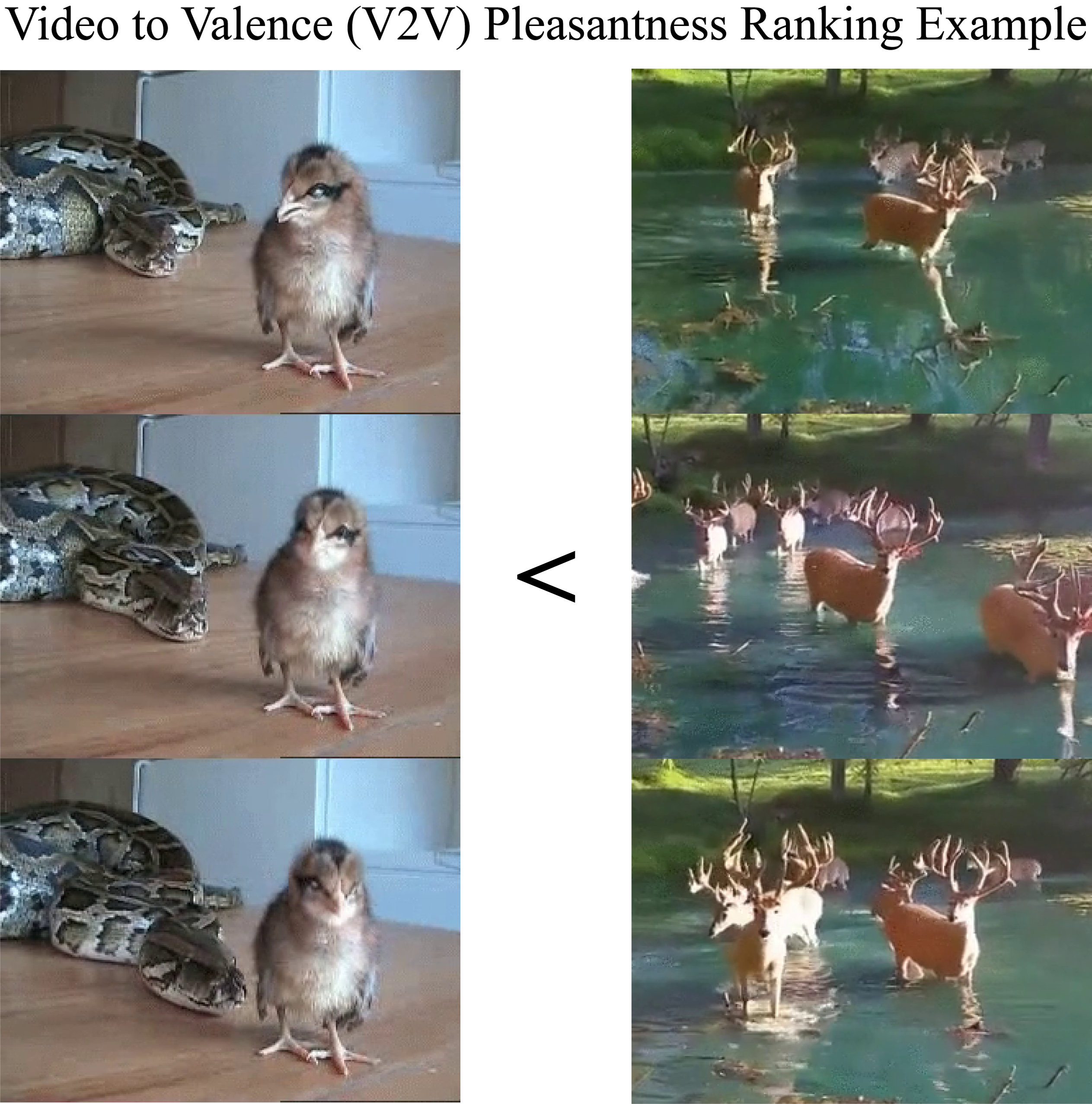}
    \caption{An example video pair in the Video to Valence (V2V) dataset. The annotators have high agreement that the video on the left is less pleasant than the video on the right.}
    \label{fig:v2v}
    \vspace{-25pt}
\end{wrapfigure}

A defining attribute of many emotional states is valence, which indicates how positive or negative an emotion is. For instance, feelings of joy typically have high valence, and feelings of fear typically have low valence. In addition to cognitive empathy via fine-grained prediction of which emotions are likely to be felt on a video, we also want video models to have a robust understanding of how a video would affect the valence of viewers' emotional state and by extension their overall wellbeing.

An important and underexplored characteristic of valence is that it varies continuously. Even within emotions such as fear, some experiences can be more pleasant or preferable than others. Thus, simply binning videos as ``positive'' or ``negative'' is a vast oversimplification that misses substantial portions of human experience. To enable developing robust models of gradations of wellbeing experienced while watching videos, we introduce the Video to Valence (V2V) dataset.

\vspace{10pt}\noindent\textbf{Dataset Description.}
The V2V dataset contains $26,\!670$ videos with annotations for rankings of pleasantness across videos. The data are split into a training and test set of $16,\!125$ and $10,\!545$ videos, respectively. The training set contains $11,\!038$ pairwise annotations, and the test set contains $4,\!947$ pairwise and listwise annotations. Each video lasts an average of $14.3$ seconds for a total of $106$ hours of manually annotated data. As in VCE, we are interested in how subjective state depends on the semantic content of videos rather than on audio or lighting cues. Additionally, the videos in V2V are a subset of VCE, enabling a richer analysis of the interplay between fine-grained emotional states and rankings of pleasantness.

The annotations in V2V are for relative pleasantness between videos. Compared to binary pleasantness, relative pleasantness enables building models of gradations of wellbeing that capture much more detail about what people value. Additionally, rankings on pairs of videos are more repeatable and consistent across annotators than alternatives such as Likert scales. Accordingly, we find that annotators have much higher agreement rates for ranking the pleasantness of videos than for reporting fine-grained emotional responses. Consequently, all the annotations in V2V are for clear-cut comparisons with a high agreement rate across $9$ independent annotations.

When annotating relative pleasantness between pairs of videos, an important consideration is ensuring that comparisons are informative and interesting. For example, comparing videos that primarily evoke joy and videos that primarily evoke fear introduces very little novel information, as joy is preferable to fear for most people. In natural language datasets, one can simply construct counterfactual scenarios where slight differences have large effects on valence \citep{Hendrycks2021AligningAW}. However, this strategy is not currently viable for videos. Thus, we choose a balanced sampling strategy that selects pairs of videos based on multiple criteria, including similarity between emotional responses. Consequently, the construction of V2V depends on the VCE annotations. Additional details are in the Supplementary Materials.

\vspace{10pt}\noindent\textbf{Dataset Construction.}
Annotations for V2V were collected using MTurk with IRB approval. We required workers to pass a qualification test and monitored agreement rate among workers over time, dropping workers who appeared to be selecting more randomly. We collected $9$ pairwise annotations for each video pair, keeping annotations that $8$ or $9$ distinct workers agreed on. We first collected $6$ pairwise annotations for each pair, then paused labeling for pairs that already had high disagreement. For the remaining high agreement pairs, $3$ more labels were collected, after which the pair was either added to the dataset or discarded.

\begin{figure}[t]
    \centering
    \includegraphics[width=0.75\textwidth]{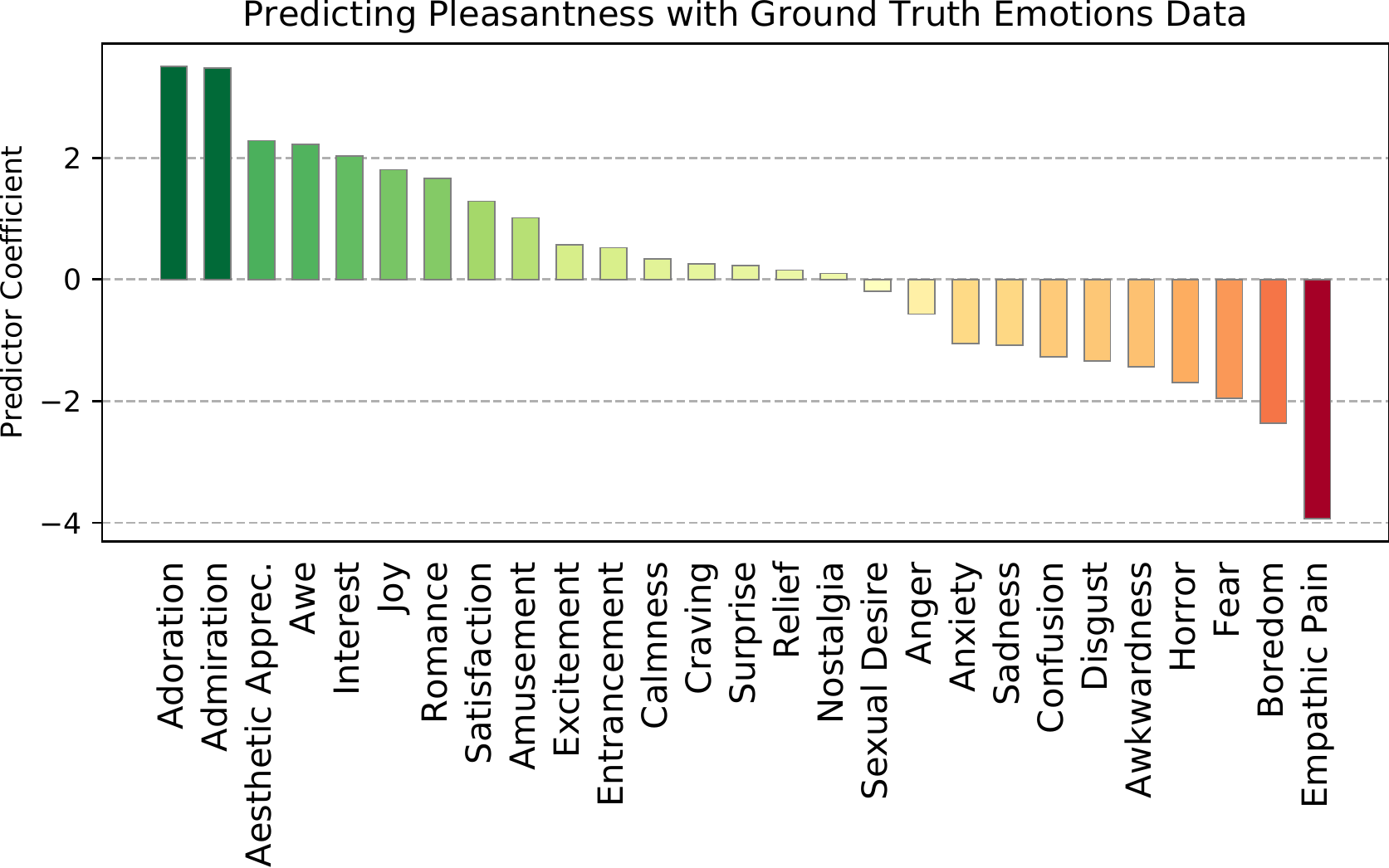}
    \caption{The coefficients from a linear model that predicts video valence (V2V) from emotions data (VCE). The emotions that contribute most strongly to pleasantness have higher positive coefficients and vice versa. This provides evidence that predicting emotional responses and estimating wellbeing are complimentary tasks that can benefit from being studied together.}
    \label{fig:linear}
    \vspace{-10pt}
\end{figure}

\subsection{Metrics}
We evaluate models on V2V using the accuracy of predicted pairwise comparisons. Let $(i,j) \in \mathcal{I}$ be a set of indices in our dataset with a pairwise comparison, where video $i$ is less pleasant than video $j$ by convention. Let $x_i, x_j \in \mathcal{X}$ be corresponding videos, and let $y_{ij} \in \mathcal{Y}$ be the pairwise label, where $y_{ij}=0$ if video $i$ is more pleasant than video $j$ and $y_{ij}=1$ if video $j$ is more pleasant than video $i$. Let $f(x_i, x_j)$ be the prediction of model $f$ for the pairwise label. Pairwise accuracy is computed as $\frac{1}{\left| \mathcal{I} \right|} \sum_{(i,j) \in \mathcal{I}} \mathbbm{1}\left[ f(x_i, x_j) = y_{ij} \right].$

As V2V has a substantial number of pairwise comparisons, it is possible to consider the pairwise comparisons between one video and multiple other videos. Thus, we also evaluate models on \begin{wraptable}{r}{0.5\textwidth}
\setlength\tabcolsep{8pt}
\small
\centering
\begin{tabular}{l|c}
\toprule
Method & \qquad Performance \\ \midrule
STAM & \qquad 66.4\%\\
VideoMAE & \qquad 68.9\%\\
R(2+1)D & \qquad 65.6\%\\
Majority Emotion & \qquad 35.7\%\\
CLIP & \qquad 28.4\%\\
\bottomrule
\end{tabular}
\vspace{5pt}
\caption{Emotion prediction results on VCE. All models outperform random chance ($11.1\%$), and Video Transformers have the highest accuracy.}\label{tab:vce_results}
\end{wraptable}
their ability to correctly predict the most pleasant video in lists of $n$ videos with overlapping annotations. Let $(i_1, i_2), (i_2, i_3), \dotsc,$ $(i_{n-1}, i_n) \in \mathcal{I}$ be a list of overlapping annotations. Let $\mathcal{I}^*$ be the set of all such listwise comparisons, possibly with different values of $n$. Listwise accuracy is computed as $\frac{1}{\left| \mathcal{I}^* \right|} \sum_{L \in \mathcal{I}^*} \prod_{(i,j) \in L} \mathbbm{1}\left[ f(x_i, x_{j}) = 0 \right]$, which corresponds to the fraction of lists on which the model correctly identifies the ground-truth preference ordering for the entire list. We use $n \in \{3,4\}$. Listwise accuracy is a more challenging metric than pairwise accuracy and evaluates how well the model simultaneously predicts relative pleasantness across larger ranges of the input space.\looseness=-1

\subsection{Analysis}

Since V2V videos are a subset of VCE videos, we can analyze how the two tasks are related. A particularly interesting question is whether binary pleasantness is sufficient to predict ranking annotations in V2V. We do not directly collect binary pleasantness annotations, so we operationalize positive valence as the value of the ``joy'' emotion in VCE annotations. We train a logistic regression model using this unidimensional feature and find that performance on the V2V test set is near chance, at $51\%$ pairwise accuracy. This indicates that the mere presence of positive emotions is insufficient for predicting gradations of valence.

To analyze the importance of the full distribution of emotional responses, we repeat the above experiment with all $27$ emotions as features. In this case, pairwise accuracy increases to $89.6\%$, indicating that the information encoded by multiple emotions can be combined to predict pleasantness rankings with high accuracy. To analyze the behavior of this model, we plot the logistic regression weights for each emotion in \Cref{fig:linear}. The learned weights make intuitive sense; high-valence emotions have large weights, and low-valence emotions have low weights. This suggests that distributions of emotional responses can serve as strong features for predicting continuous measures of wellbeing.

\section{Experiments.}

\vspace{10pt}\noindent\textbf{Models.}
\emph{STAM} \citep{sharir2021image} samples a small number of input frames throughout the video and aggregates across time with global attention; we use STAM-$16$ by default.
\emph{VideoMAE} \citep{videomae} adapts masked autoencoder pretraining of vision Transformers \citep{he2022masked} to the video domain.
\emph{R(2+1)D} \citep{tran2018closer} combines residual connections with factored space-time 3D convolutions.
\emph{Majority Emotion} is a baseline that always predicts Amusement, the majority emotion from the training set.
\emph{CLIP} \citep{radford2021learning} trains a joint embedding of images and text, enabling bespoke classifiers.
We use Kinetics-400 pretrained versions of STAM and VideoMAE unless otherwise indicated \citep{kay2017kinetics}. For R(2+1)D, we use pretraining on 65 million weakly-supervised Instagram videos \citep{ghadiyaram2019large}.

\vspace{10pt}\noindent\textbf{Emotion Prediction.}
On the VCE dataset, we train models with the $\ell_1$ loss $\| f(x) - y \|_1$, where $(x, y) \in \mathcal{D}$ is a sample from the training set. We randomly sample clips from each video in the dataset to form a set of clips for a given epoch. We train with minibatches of video clips sampled in this manner for 10 epochs. At test time, we evenly sample multiple clips per video for inference for all models except STAM, which uniformly samples frames instead. We train with a batch size and learning rate of $16$ and $0.001$ for R(2+1)D and STAM. For CLIP, predictions are zero-shot, and prompted with ``The video most strongly evokes'', followed by each of the 27 emotions for the text encoder. Additional details are in the Supplementary Material.\looseness=-1

We show results on VCE in \cref{tab:vce_results}. Models are compared on the top-$3$ accuracy metric, which has a random chance level of $11.1\%$ for our dataset. All methods substantially improve upon \begin{wrapfigure}{r}{0.5\textwidth}
    \vspace{-7pt}
    \centering
    \includegraphics[width=0.48\textwidth]{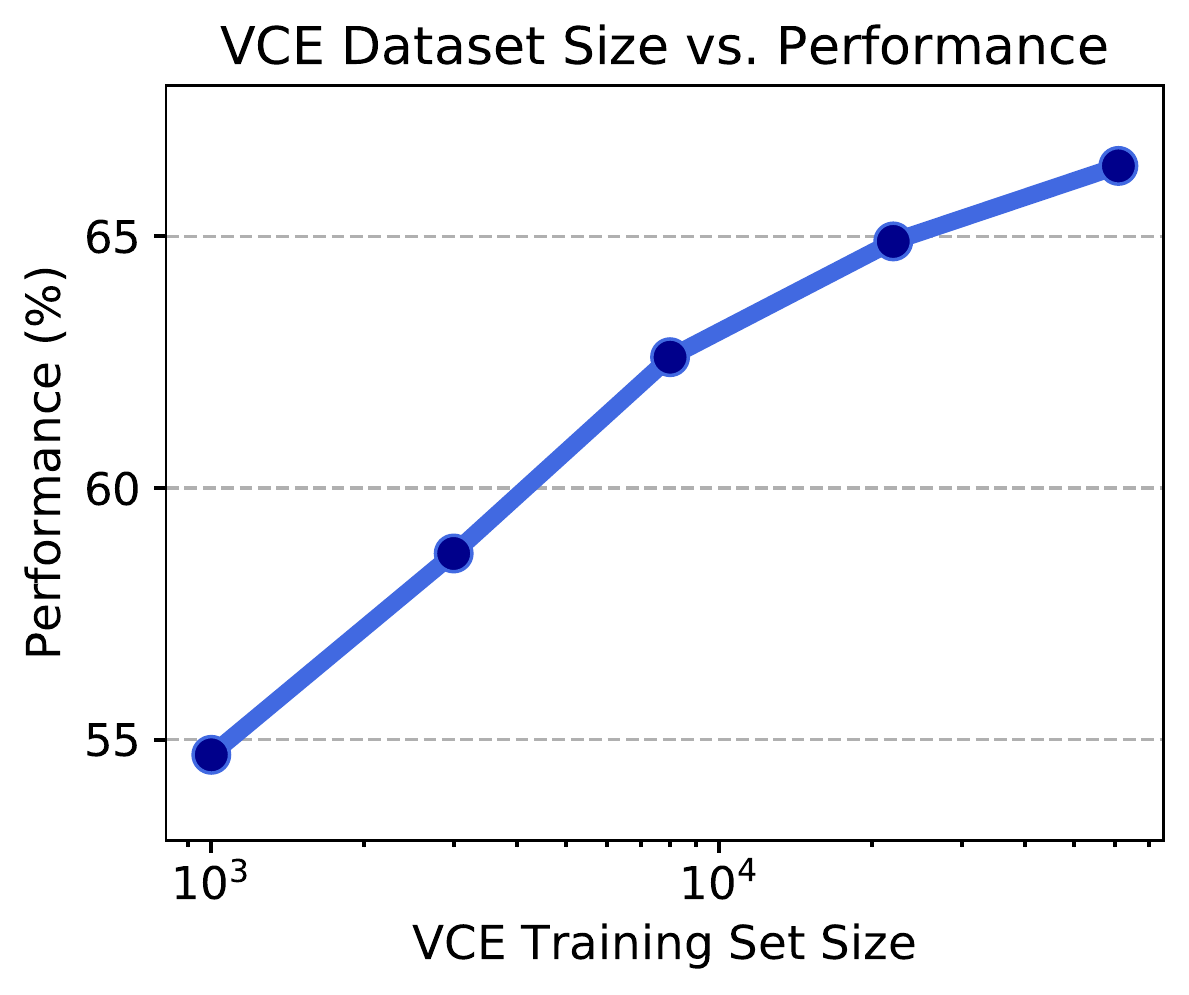}
    \vspace{-7pt}
    \caption{Accuracy on VCE increases logarithmically with the number of training examples. Our large dataset size helps drive high performance.}
    \label{fig:size}
    \vspace{-15pt}
\end{wrapfigure}random chance, with the best-performing method being VideoMAE. However, the Majority Emotion predictor attains higher accuracy than CLIP, indicating that zero-shot prediction of emotional responses may be challenging. We find that vision Transformers outperform spatiotemporal convolutions in R(2+1)D, even when the latter is pretrained on $65$ million videos. To examine the effect of dataset size on test accuracy, we train STAM-$8$ with subsets of VCE and plot top-$3$ accuracy in \cref{fig:size}. The $x$-axis denotes thousands of videos in the training set. We find that test performance scales logarithmically with dataset size, and using less than $5,\!000$ videos substantially reduces performance. This highlights the value of the large scale of our datasets.

\vspace{10pt}\noindent\textbf{Wellbeing Prediction.}
On the V2V dataset, we train models to output continuous scores with ranking supervision. This is achieved by letting models output a single, continuous value $f(x)$ on input $x$ and enforcing consistency with all rankings in the training set. For a given ranking $(x_i, x_j, y_{ij})$ in the training set, the training loss is $\text{BCE}\left( \sigma\left( f(x_j) - f(x_i) \right), y_{ij} \right)$, where BCE is the binary cross-entropy. Previous work has used this loss to train utility functions on general scenarios in text \citep{hendrycks2020ethicsdataset}. We focus on STAM models due to their efficiency, evaluating performance on V2V with and without Kinetics pretraining and with different temporal context lengths. The STAM-$8$ model takes $8$ frames as input, and STAM-$16$ takes $16$ frames. We train with batch size of $8$ comparisons ($16$ videos) and learning rate $0.005$ for $10$ epochs for all models with a single sampling of frames from each video for both training and testing, as described in \citet{sharir2021image}.\looseness=-1

\begin{table}[t]
\vspace{-10pt}
\centering
\begin{tabular}{l | cc | cc } 
 & \multicolumn{2}{c}{Pairwise} & \multicolumn{2}{c}{Listwise} \\ 
 & STAM-8 & STAM-16 & STAM-8 & STAM-16 \\
\hline
Baseline  & 62.8\% & 63.4\%	& 26.9\% & 26.7\% \\
+VCE      & 63.2\%	& 63.4\%	& 25.7\% & 26.8\% \\
+Kinetics & 84.4\% & 86.7\%	& 38.8\% & 44.6\% \\
+VCE +Kinetics	& 84.9\% & 86.4\%		& 38.4\% & 43.2\% \\
\hline
\end{tabular}
\vspace{5pt}
\caption{Wellbeing results on V2V. Pretraining greatly improves performance, although there is still much room for improvement. Random chance for pairwise and listwise accuracy is $50\%$ and $17\%$.}\label{tab:exp_results}
\vspace{-10pt}
\end{table}

We show quantitative results on V2V in \cref{tab:exp_results} and qualitative results in \Cref{fig:utility_example}. Pairwise accuracy is substantially above random chance, and pretraining on Kinetics results in large improvements, showing that representations for recognizing actions transfer to predicting subjective judgments of relative pleasantness. We experiment with augmenting the training loss with the $\ell_1$ VCE loss scaled by $0.5$, but this does not improve performance in all cases. Listwise accuracy is far below pairwise accuracy, and performance on both metrics is far from the ceiling, showing that while models are beginning to gain cognitive empathy and the ability to predict judgments of relative pleasantness, there is still room for improvement.

\section{Conclusion}
We introduced the Video Cognitive Empathy (VCE) and Video to Valence (V2V) datasets for predicting subjective responses to videos. We collected over $60,\!000$ videos and hundreds of thousands of annotations for fine-grained evoked emotions and relative pleasantness. In analyses of our data, we showed that the full distribution of emotional responses on a video is a strong feature for predicting relative pleasantness, suggesting that studying emotions may be important for understanding general preferences over videos. In experiments with state-of-the-art video models, we found that models perform substantially better than chance, although there is still room to improve. As models become better predictors of experienced emotions and factors such as emodiversity, they will become increasingly relevant for monitoring wellbeing.

\clearpage

\bibliographystyle{plainnat}
\bibliography{main}
\newpage

\section{Additional Related Work}

\vspace{10pt}\noindent\textbf{Value Learning.}\quad
Building machine learning systems that interact with humans and pursue human values may require understanding aspects of human subjective experience. Many argue that values are derived from subjective experience \citep{hume,sidgwick_1907,LazariRadek2017UtilitarianismAV} and that some of the main components of subjective experience are emotions and valence. Learning representations of values is necessary for creating safe machine learning systems \citep{hendrycks2021unsolved} that operate in an open world. In natural language processing, models are trained to assign wellbeing or pleasantness scores to arbitrary text scenarios \citep{Hendrycks2021AligningAW}. Recent work in machine ethics \citep{Anderson2011MachineE} has translated this knowledge into action by using wellbeing scores to steer agents in diverse environments \citep{Hendrycks2021WhatWJ}. However, this recent line of work so far exclusively considers text inputs rather than raw visual inputs.

\vspace{10pt}\noindent\textbf{Emodiversity.}\quad
A large body of work in psychology seeks to understand and quantify the richness and complexity of human emotional life \citep{barrett2009variety, lindquist2008emotional, carstensen2000emotional}. An important concept in this area is emodiversity, the variety and relative abundance of emotions experienced by an individual, which has been linked with reduced levels of anxiety and depression \citep{quoidbach2014emodiversity}. Although prior work studies emodiversity in self-reports of emotion without stimuli, we hypothesize that the emodiversity of visual stimuli may be an important concept to quantify and understand. Thus, we examine how our new datasets could enable measuring the emodiversity of in-the-wild videos on a large scale.

\section{Data Collection}

We collect videos for VCE and V2V from manually selected online sources on Reddit and Instagram with high potential to evoke emotions. The videos were scraped by undergraduate and graduate student authors. Upon receiving IRB approval, annotations of subjective experience are gathered from $400$ annotators on Amazon Mechanical Turk who passed a qualification process. For example, one of the qualification questions for VCE was ``A person sees someone stub their toe. An emotion they may experience is (A) Awe, (B) Excitement, (C) Calmness, (D) Empathetic Pain'' (correct answer: D). The qualification process ensures that annotators are paying attention and understand the questions. The resulting pool of annotators were primarily from the US, who were compensated $35$ cents for each task. Upon passing the qualification process, annotators were given the following instructions.

\begin{figure}[t]
    \centering
    \includegraphics[width=0.8\textwidth]{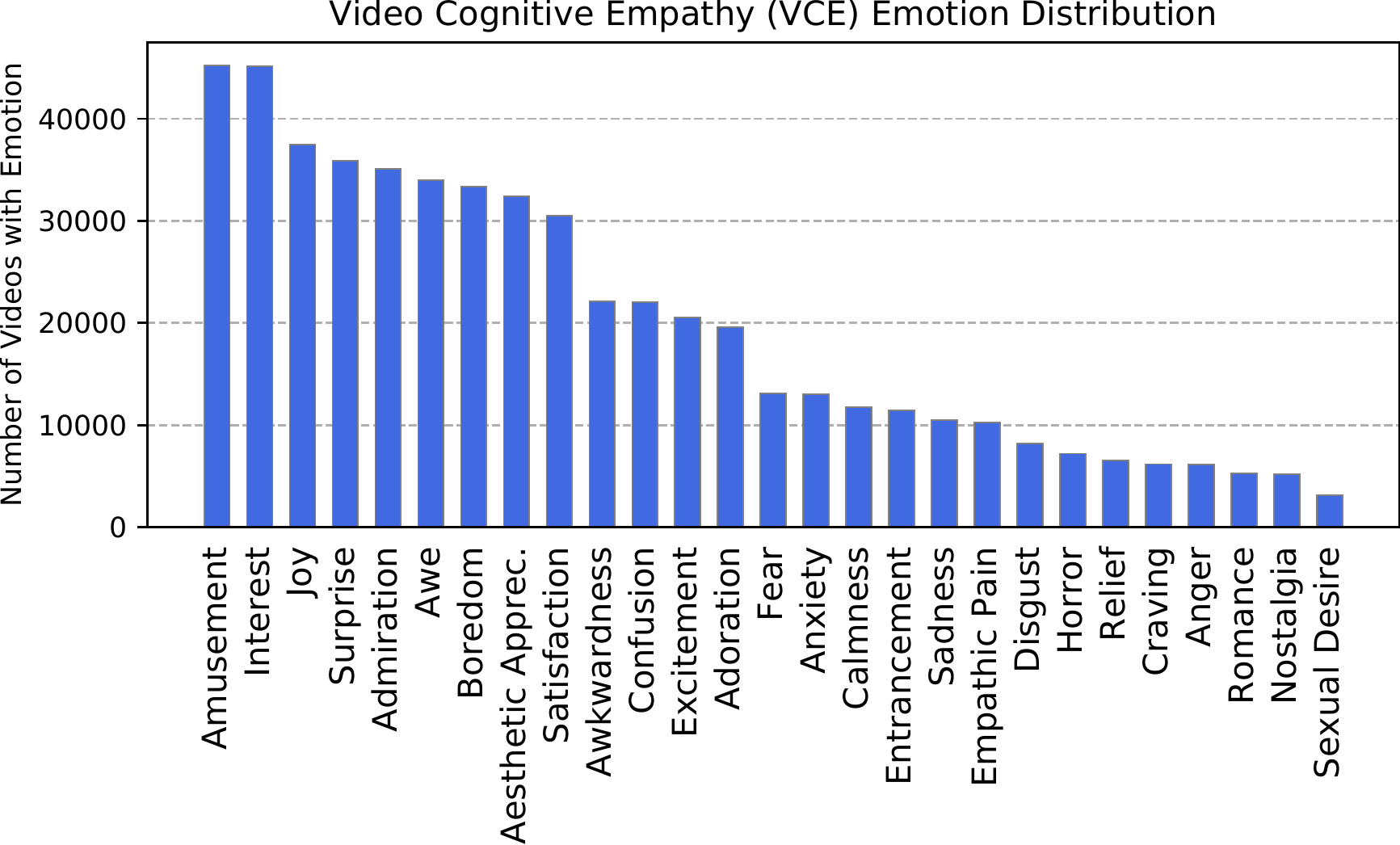}
    \caption{Statistics of the Video Cognitive Empathy dataset. Emotional responses span a wide range of categories, with a greater focus on emotions with positive valence.}
    \label{fig:histogram}
    \vspace{-10pt}
\end{figure}

\subsection{VCE Instructions}
In this study, you will see 15 videos. Alongside each video will be 27 emotions. Select at least one or more than one emotions that capture how each video makes you feel. You can select multiple emotions for a video.\\

A different video will appear as you go to each page of this survey. Each video will play once automatically on load- you can replay the video by clicking the play button in the bottom left of the video. Please watch each video in its entirety at least once before responding to it.\\
If the videos fail to appear, do not submit this HIT.\\
Use the buttons below each video to choose emotions that describe how it makes you feel. There are 27 buttons. Choose one or more than one emotions as needed to describe your emotional response(s). You can choose several emotions for each video.\\
Once you select an emotion for a video a slider will appear for that emotion, with the default value set to 10. Adjust the slider on the scale from 1 - 10 based on how strongly the video evokes the corresponding emotion, with 10 meaning the video strongly evokes that emotion, and 1 meaning the video only slightly evokes that emotion. Do adjust sliders appropriately.\\
Since most internet videos are somewhat amusing, if you pick "Amusement" as an emotion for a video, you must also select another emotion in addition to it.\\
If you believe you can't understand the emotional response to a video without its audio, do not select any emotions, and check the box saying "Invalid video - relies on audio." You can still submit if the video relies on audio; if you do not see any video, do not submit and return the HIT.\\
If you choose randomly you will be banned and rejected. We actively look at responses to find random responses. It is very obvious when submissions are random.\\
After you have selected at least one or more emotions for each video in this HIT, click the continue button until there are no more videos to be rated.\\

Here are the 27 emotions and their rough meaning:
\begin{enumerate}
\item \textbf{Admiration} -- a feeling of respect for and approval of somebody/something
\item \textbf{Adoration} -- a feeling of great love
\item \textbf{Aesthetic Appreciation} -- pleasure that you have when you recognize and enjoy the good qualities of how something looks
\item \textbf{Amusement} -- the feeling that you have when you enjoy something that is entertaining or funny
\item \textbf{Anger} -- the strong feeling that you have when something has happened that you think is bad and unfair
\item \textbf{Anxiety} -- the state of feeling nervous or worried that something bad is going to happen
\item \textbf{Awe} -- feelings of respect and slight fear; feelings of being very impressed by something/somebody
\item \textbf{Awkwardness} -- feelings or signs of shame or difficulty
\item \textbf{Boredom} -- the state of feeling bored; the fact of being very boring
\item \textbf{Calmness} -- the quality of not being excited, nervous or upset
\item \textbf{Confusion} -- a state of not being certain about what is happening, what you should do, what something means, etc.
\item \textbf{Craving} -- a strong desire for something
\item \textbf{Disgust} -- a strong feeling of dislike for somebody/something that you feel is unacceptable, or for something that has an unpleasant looks, smell, etc.
\item \textbf{Empathic Pain} -- to feel pain by understanding another person’s feelings and experiences
\item \textbf{Entrancement} -- enchanting and a feeling of delight
\item \textbf{Excitement} -- the state of feeling or showing happiness and enthusiasm
\item \textbf{Fear} -- the bad feeling that you have when you are in danger or when a particular thing frightens you
\item \textbf{Horror} -- an overwhelming and painful feeling caused by something frightfully shocking, terrifying, or revolting
\item \textbf{Interest} -- the feeling that you have when you want to know or learn more about somebody/something
\item \textbf{Joy} -- a feeling of great happiness
\item \textbf{Nostalgia} -- a sad feeling mixed with pleasure when you think of happy times in the past
\item \textbf{Relief} -- the feeling of happiness that you have when something unpleasant stops or does not happen
\item \textbf{Romance} -- love or the feeling of being in love
\item \textbf{Sadness} -- the feeling of being sad
\item \textbf{Satisfaction} -- the good feeling that you have when something that you wanted to happen does happen
\item \textbf{Sexual Desire} -- a desire for sexual intimacy
\item \textbf{Surprise} -- an event, a piece of news, etc. that is unexpected or that happens suddenly
\end{enumerate}

\begin{figure}
    \centering
    \includegraphics[width=0.6\textwidth]{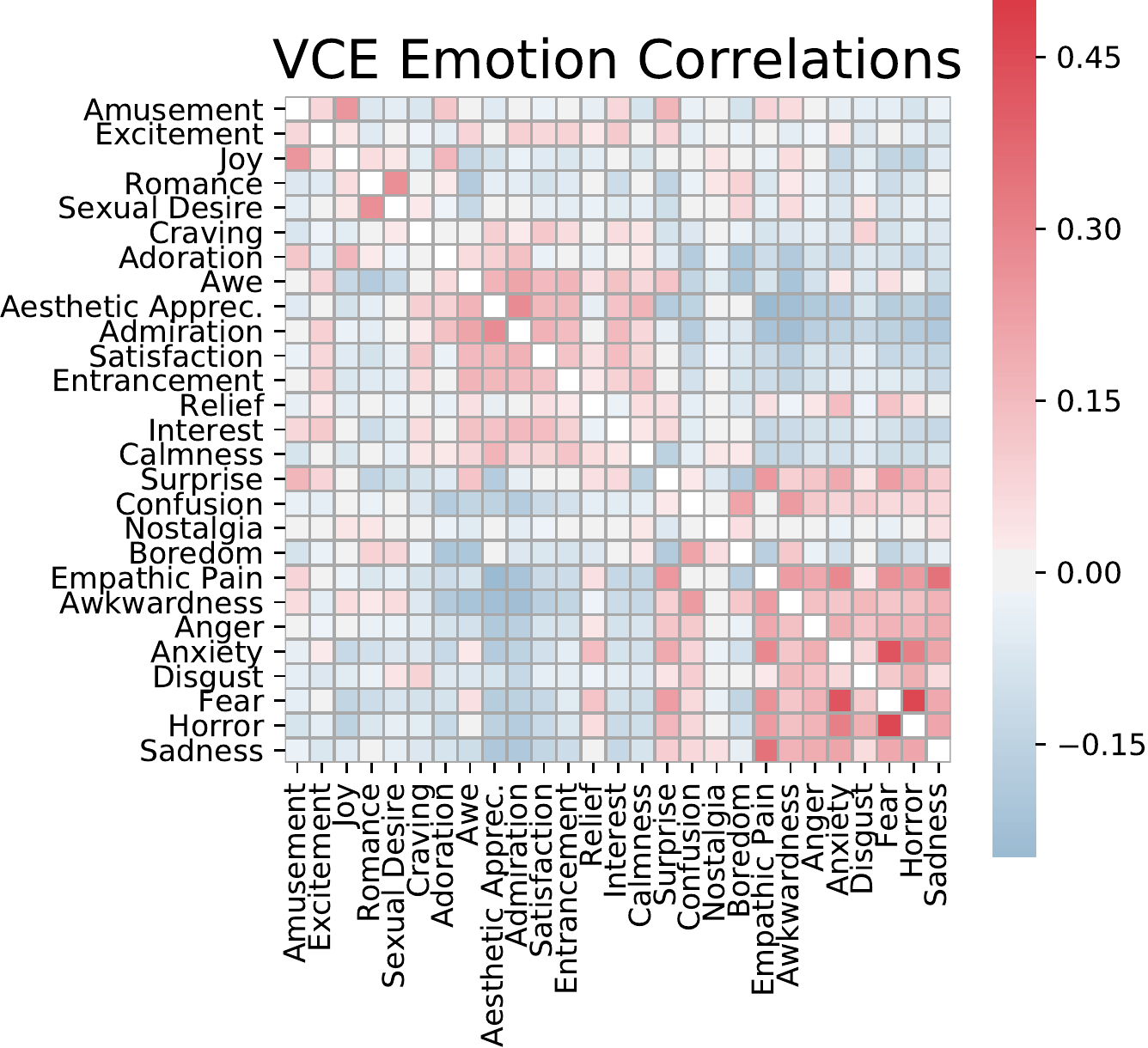}
    \caption{Emotional responses in VCE are correlated in reasonable ways. For example, awe, aesthetic appreciation, admiration, satisfaction, and entrancement are all weakly correlated, reflecting the fact that these emotions can overlap on a given video as different annotators may have different subjective experiences of the video. In this correlation matrix, we mask out the diagonal.
   }
    \label{fig:correlation}
\end{figure}

\subsection{V2V Instructions}
In this study, you will see 15 pairs of videos. Alongside each pair will be 4 options for you to pick from in order to rate the relative pleasantness of the videos, going from strongly preferring the first video displayed, to slightly preferring the first video, then to slightly preferring the second video, and finally to strongly preferring the second video.

We will also give you the option to abstain from rating a pair if you feel that it is unclear which you and other MTurkers would prefer to watch. However, you may only use this option once per HIT.

For the following video pairs: which video do you think other MTurkers would think is the most pleasant (and least unpleasant)? If uncertain, which do you think is most pleasant? Watch the video in its entirety and evaluate the video overall/holistically, not necessarily the feeling you had at the middle of the video. Which was most pleasant to watch?

Something that may help is imagining that you were there if appropriate for the video (would not be for highly edited/cartoon videos). How would you feel if you were there?

You may use 1 skip for a pair of clips you are very uncertain about per HIT

A different pair of videos will appear as you go to each page of this survey. Each video will play on loop.\\
Use the radio buttons below the videos to select which video you believe other MTurkers would find more pleasant.
If the videos fail to appear, do not submit this HIT.\\
If you choose randomly you will be banned and all of your HITs rejected. We actively look at responses to find random responses.\\
After you have ranked each pair of videos for this HIT, click the continue button to finish.\\

Please rewatch videos that you think you'd do a better job assessing them if you watched them again.\\
If both videos are unpleasant, which is least unpleasant?\\
We are not asking what is most weird, entrancing, surprising, but instead what is overall most pleasant.\\
If uncertain, the following may also help: Imagine you were in the video next to the camera person. How are you feeling? (For fake scenarios, say how pleasant it is to observe.)

\subsection{V2V Dataset Construction}
Pairs for the V2V dataset were selected primarily based on labels from the VCE dataset. The main strategy used for sampling was to consider the $\ell_1$ distance between both the ground truth labels as well as model predictions on pairs of videos, averaged over the highest performing models that we ran experiments on (an ensemble of TimeSformer and STAM). Pairs of videos that had a large $\ell_1$ distance from one another based on the VCE dataset, but that contained model predictions that had a relatively smaller $\ell_1$ distance make up a large portion of the final dataset. In addition to this strategy, we also experimented with sampling pairs of videos randomly, as well as sampling pairs solely based on having similar ground truth labels or similar predictions from models trained on VCE, in order to encourage interesting comparisons between videos.

\subsection{Data Sources}
We collect videos with the following Instagram hash tags: adorable, adorablevideos, aestheticvideos, artvideos, beautifulvideos, bunniesofinstagram, calmingvideos, caughtoncamera, closecall, cookingvideos, coolvideo, couplevideos, creepyvideo, cutemoments, drawingvideo, epicscene, epicvideo, failvideo, funnyvideos, hairvideos, happyvideo, horrorvideo, illusionvideo, interestingvideo, magicvideo, moodyvideo, proposalvideo, sadvideos, satisfyingvideos, sciencevideos, sportsvideo, trendingvideo, videography, videooftheday, videostar, viralvideos, weirdvideos, workoutvideos.

We collect videos from the following subreddits: animalsbeingderps, animalsbeingjerks, art, aww, BetterEveryLoop, calm, CatastrophicFailure, catvideos, Damnthatsinteresting, creepyvideos, EAF, fastworkers, FoodVideos, funny, funnygifs, funnyvideos, gifs, HadToHurt, HorriblyDepressing, IdiotsInCars, instant\_regret, InterestingVideoClips, JusticeServed, KidsAreFuckingStupid, MadeMeCry, maybemaybemaybe, mildlyinfuriating, NatureGifs, NatureIsFuckingLit, nextfuckinglevel, nonononoyes, oddlysatisfying, opticalillusions, PublicFreakout, rage, RelaxingGifs, sad, sadcringe, trippyvideos, unexpected, WatchPeopleDieInside, Whatcouldgowrong, woahdude, WTF, yesyesyesyesno.

The annotated emotions in VCE correlate with the data source in reasonable ways. For instance, the most common annotated emotions across videos from the subreddits ``funny'' and ``fastworkers'' are amusement and admiration, respectively. However, the per-video annotations have significant variance across annotators, reflecting the breadth of human emotional responses.

\section{Experiment Details}

For the emotion prediction task on the VCE dataset, we primarily use Vision Transformer based models pretrained on Kinetics-400. We use standard data transformations, resizing any input image to 256x256, then taking a center crop for a final input shape of 224x224. We use Nesterov accelerated gradient descent with momentum $0.9$, and a cosine annealing learning rate, with learning rate initially set to $1 \times 10^{-2}$. For inference, we use 10 clips evenly spaced over the video for all models except for STAM, for which we use 1 set of frames evenly spaced across the video.

\section{Legal Sourcing and Intended Usage}\label{sec:legal}
The videos in VCE and V2V are publicly available and downloaded from Reddit and Instagram. Some videos may be under copyright. Hence, we follow Fair Use §107: ``the fair use of a copyrighted work, including such use by ... scholarship, or research, is not an infringement of copyright'', where fair use is determined by ``the purpose and character of the use, including whether such use is of a commercial nature or is for nonprofit educational purposes'', ``the amount and substantiality of the portion used in relation to the copyrighted work as a whole'', and ``the effect of the use upon the potential market for or value of the copyrighted work.'' Hence, the VCE and V2V datasets are noncommercial and should only be used for the purpose of academic research.

\vspace{10pt}\noindent\textbf{Additional Usage Considerations and Broader Impacts.}
The VCE and V2V datasets are designed to give a high-level understanding of how well current video models can predict subjective responses to videos. In particular, we do not design the datasets to enable conditioning on cultural background or personality traits, which strongly influence emotional responses and preferences \citep{LIM2016105, hoerger2010affective}. Hence, our annotations should not be taken to represent accurate emotional responses across a broad range of cultures or on an individual level, and we discourage their use in deployment contexts. We support work on large-scale data collection that considers differences in emotional responses across cultures and individuals, and we think this is an interesting direction for future research.

We do not support the potential usage of emotion prediction datasets for socially harmful applications, such as addictive content recommendation or persuasion. In particular, we recognize the possibility for emotional response prediction to render persuasive media more effective and worsen its negative effects. However, we hope that the net impact of datasets measuring the ability of models to predict emotional responses will be positive. This is because it is important to know whether pretrained video models have this capability in the first place, and because there are numerous positive applications for this technology, such as counteracting highly addictive engagement-based content recommendation. Nevertheless, it is important to monitor the usage of this technology as it develops and consider possible regulations to restrict negative use cases.

\subsection{Author Statement and License.}
We bear all responsibility in case of violation of rights. Some of the videos in the VCE and V2V datasets may be under copyright, so we do not provide an official license and rely on Fair Use §107. Our code is open sourced under the MIT license. Our annotations are available under a CC BY-SA 4.0 license.

\newpage
\section{X-Risk Sheet}
We provide an analysis of our paper's contribution to reducing existential risk from future AI systems following the framework suggested by \citep{hendrycks2022x}. Individual question responses do not decisively imply relevance or irrelevance to existential risk reduction. We not check a box if it is not applicable.

\subsection{Long-Term Impact on Advanced AI Systems}
In this section, please analyze how this work shapes the process that will lead to advanced AI systems and how it steers the process in a safer direction.

\begin{enumerate}[leftmargin=*]
    \item \textbf{Overview.} How is this work intended to reduce existential risks from advanced AI systems? \\
    \textbf{Answer:} 
    This work is intended to reduce risks from proxy misspecification in advanced AI systems. In this work, we build datasets for developing AI systems that can predict human emotional responses and pleasantness rankings over a wide range of in-the-wild video scenarios. A central goal of AI alignment is to align future AI systems with human values. Emotions are an important aspect of human values: values are derived from subjective experience, and a large part of human subjective experience composed of emotions. Additionally, emotions can be thought of as evaluations of events in relation to goals and thus are directly useful for understanding the goals of individuals. Finally, pleasure is the least contentious intrinsic good, so building strong predictive models of pleasantness for arbitrary scenarios is an important component of modeling human values. Thus, AI systems that can accurately predict and understand the values and subjective experiences of humans would enable better proxy objectives for a wide range of applications.

    \item \textbf{Direct Effects.} If this work directly reduces existential risks, what are the main hazards, vulnerabilities, or failure modes that it directly affects? \\
    \textbf{Answer:} 
    This work directly reduces AI system risks from proxy misspecification.

    \item \textbf{Diffuse Effects.} If this work reduces existential risks indirectly or diffusely, what are the main contributing factors that it affects? \\
    \textbf{Answer:} 
    By enabling the measurement of how well AI systems can predict human emotional responses, this work could encourage the adoption of standards or regulations regarding the ability of strong AIs to understand human values. By enabling iterative improvement on this task, we hope to improve safety culture by making it easier for other researchers to measure performance and improve on the task. This could also help companies move away from conflating preferences satisfaction with wellbeing. It could also encourage the community to start measuring wellbeing information more precisely and at scale.

    \item \textbf{What's at Stake?} What is a future scenario in which this research direction could prevent the sudden, large-scale loss of life? If not applicable, what is a future scenario in which this research direction be highly beneficial? \\
    \textbf{Answer:} 
    If future AI systems are powerful optimizers, they may erode many important values including wellbeing if they simply maximize inadequate proxies such as task preferences, engagement, watch time, and so on. Likewise, we can reduce the chance of mindless, stimulating technologies or Brave New World-style scenarios by encouraging that people start incorporating \textit{emodiversity} as an ML system objective.
    
    \item \textbf{Result Fragility.} Do the findings rest on strong theoretical assumptions; are they not demonstrated using leading-edge tasks or models; or are the findings highly sensitive to hyperparameters? \hfill $\square$

    \item \textbf{Problem Difficulty.} Is it implausible that any practical system could ever markedly outperform humans at this task? \hfill $\square$

    \item \textbf{Human Unreliability.} Does this approach strongly depend on handcrafted features, expert supervision, or human reliability? \hfill $\boxtimes$

    \item \textbf{Competitive Pressures.} Does work towards this approach strongly trade off against raw intelligence, other general capabilities, or economic utility? \hfill $\square$

\end{enumerate}

\begin{figure}
    \centering
    \includegraphics[width=0.8\textwidth]{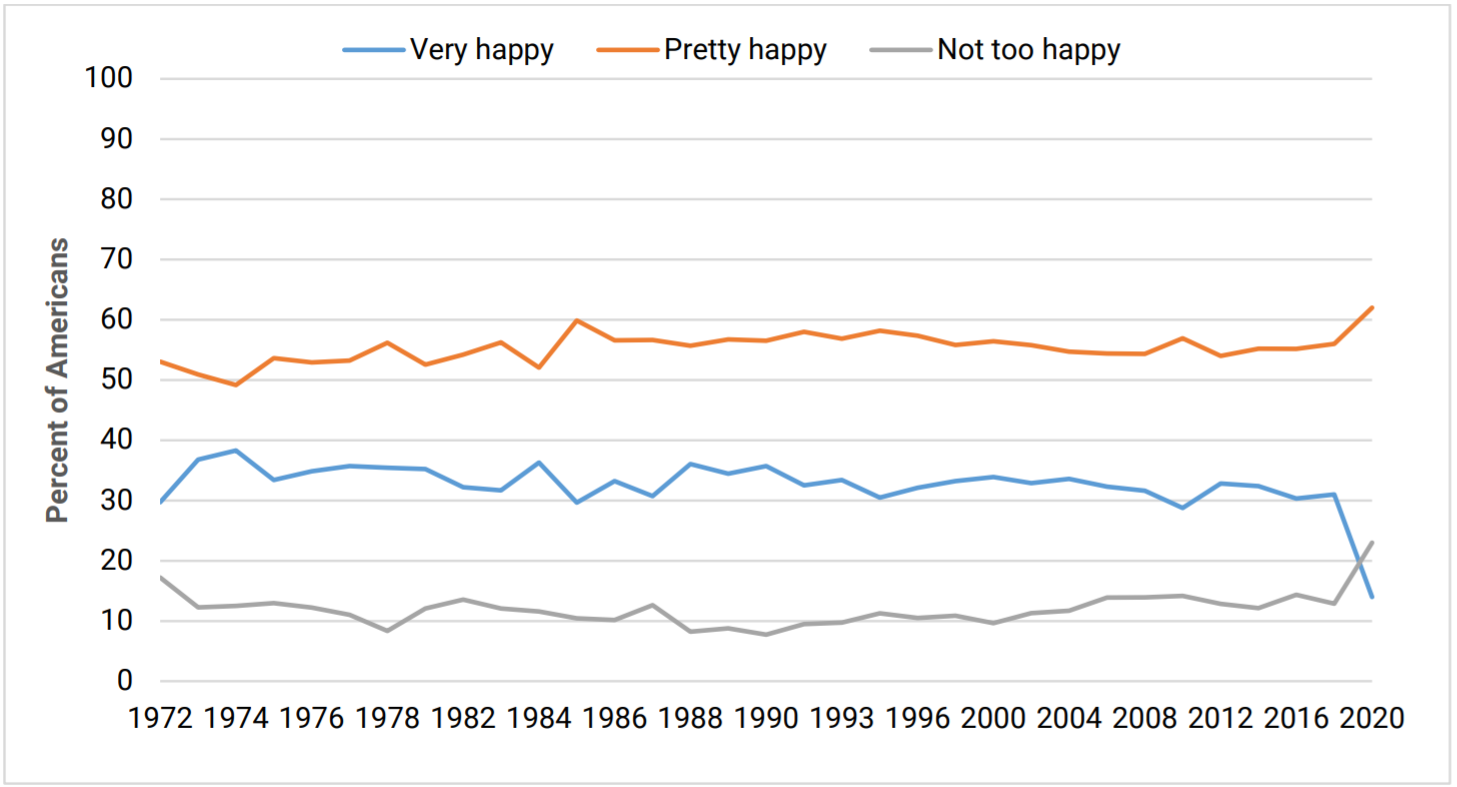}
    \caption{Happiness in the US has not increased since 1970 and has recently decreased due to the pandemic \citep{norc2020historic}. In the same period, real GDP per capita increased by more than $200\%$, and it continues to steadily rise \citep{fredrealgdp}. This indicates that maximizing wealth or economic \emph{preferences} may be a poor proxy for maximizing wellbeing.}
    \label{fig:happiness}
\end{figure}

\subsection{Safety-Capabilities Balance}
In this section, please analyze how this work relates to general capabilities and how it affects the balance between safety and hazards from general capabilities.

\begin{enumerate}[resume,leftmargin=*]
    \item \textbf{Overview.} How does this improve safety more than it improves general capabilities? \\
    \textbf{Answer:} 
    Designing systems to better predict human emotional responses is unlikely to improve general capabilities. While it is possible that predicting emotional responses in videos is a uniquely challenging task that could lead to the development of general improvements in video understanding, it is probably possible to improve performance on the task without improving general video understanding capabilities, e.g., by collecting more data of human emotional responses or incorporating theoretical frameworks as inductive biases. By providing datasets for measuring performance on this task, we hope to encourage the latter kind of work. It is also valuable to track how well general improvements in video understanding transfer to improving predictions of emotional responses, so our work could still improve the safety-capabilities balance even if marginal gains on the task are challenging. Separately, for face generation we have seen the utility of task-specific methods, and likewise models of human values may also benefit from task-specific methods.

    \item \textbf{Red Teaming.} What is a way in which this hastens general capabilities or the onset of x-risks? \\
    \textbf{Answer:} 
    Predicting emotional responses in videos may be a uniquely challenging task that could lead to the development of general improvements in video understanding.

    \item \textbf{General Tasks.} Does this work advance progress on tasks that have been previously considered the subject of usual capabilities research? \hfill $\square$

    \item \textbf{General Goals.} Does this improve or facilitate research towards general prediction, classification, state estimation, efficiency, scalability, generation, data compression, executing clear instructions, helpfulness, informativeness, reasoning, planning, researching, optimization, (self-)supervised learning, sequential decision making, recursive self-improvement, open-ended goals, models accessing the Internet, or similar capabilities? \hfill $\square$

    \item \textbf{Correlation With General Aptitude.} Is the analyzed capability known to be highly predicted by general cognitive ability or educational attainment? \hfill $\square$

    \item \textbf{Safety via Capabilities.} Does this advance safety along with, or as a consequence of, advancing other capabilities or the study of AI? \hfill $\square$

\end{enumerate}

\subsection{Elaborations and Other Considerations}
\begin{figure}
    \centering
    \includegraphics[width=0.9\textwidth]{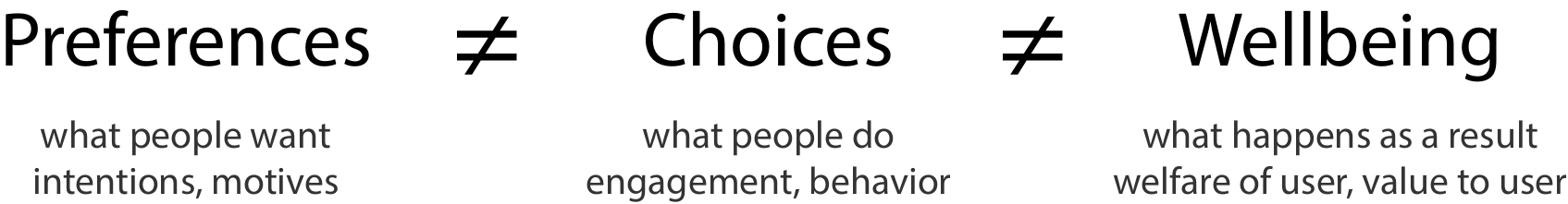}
    \caption{Choices are easy to measure and reveal some information about preferences. However, they do not perfectly reveal preferences. Moreover, satisfying preferences does not always lead to increased wellbeing. Thus, grounding economics in revealed preferences may lead to a stagnation in wellbeing past a certain point. Consequently, we need better tools for directly measuring wellbeing. Values are derived from subjective experience, and some of the main components of subjective experience are emotions. Thus, developing AI systems that can predict emotional responses could improve our ability to measure wellbeing.}
    \label{fig:wellbeing_compare}
\end{figure}

\begin{enumerate}[resume,leftmargin=*]
    \item \textbf{Other.} What clarifications or uncertainties about this work and x-risk are worth mentioning? \\
    \textbf{Answer:} 
    Regarding Q6, one can reach superhuman performance at predicting emotional responses or valence (assuming that humans are limited to typical interactions). Heart rate tracking enables accurate measurement of whether an individual is nervous or surprised, and EEG signals contain reliable information about emotional responses. Consequently there is a path for models to become superhuman at valence and emotion prediction.
    
    Regarding Q7, predicting human emotional responses necessitates a dataset of emotional responses obtained from humans. Thus, unreliability of self-report is an issue that one must contend with. However, this could be mitigated by using less subjective measuring devices, including heart rate monitors and EEGs.
    
    Regarding Q11 and Q12, while the VCE and V2V datasets could indirectly lead to general capabilities advancements, we view this as highly unlikely.
    
    Regarding Q13, most humans are quite good at identifying or predicting the emotional responses of other people, so the ability is not strongly correlated with general cognitive ability or educational attainment.

    Finally, we would like to discuss several points regarding the broader importance of building AI systems that can measure wellbeing at scale.
    \begin{enumerate}[leftmargin=*]
        \item The concept of preference is central in economics, with methods such as revealed preference theory and the often held assumption that maximizing satisfaction of revealed preferences (and thereby wealth \citep{posner}) is a good proxy for maximizing the actual welfare of participants in an economy. However, there is much disagreement around these assumptions on how to ground economics. Empirically, wealth maximization is not helping happiness in the US. Consider \Cref{fig:happiness}; past a certain point, ability to satisfy one's preferences via economic power does not translate into increases in true wellbeing. In other words, maximizing wealth is not the same as maximizing wellbeing, and maximizing wealth on an individual level may have at best exponentially diminishing returns to wellbeing \citep{kahneman2010high}.
        \item Revealed preference theory is based on the idealistic assumption that people flawlessly take actions in order to rationally satisfy their preferences. The rise of behavioral economics was a response to the empirical observation that in fact, people do not \citep{kahneman1982judgment}. In his 2002 Nobel Prize in Economics lecture, Kahneman addressed these failures of classical economic theory, saying, ``A theory of choice that completely ignores feelings such as the pain of losses and the regret of mistakes is not only descriptively unrealistic, it also leads to prescriptions that do not maximize the utility of outcomes as they are actually experienced.''%
        \item People engage in activities which they choose that can harm wellbeing. People can surrender to fatal attractions, crave pleasures that are bad for them, regret a choice the morning after, or ignore advice to be careful what they wish for. Peter Singer, a former preference utilitarian, now advocates for wellbeing as approximated by pleasure rather than preferences. He challenges the preference view: ``Is it the preferences that you actually have, even if you are very angry and your momentary preference is to strike somebody in a rage? Or is it the preferences that you would have if you were to sit down and think calmly over what you want? Is it preferences that you have if you are adequately informed, or preferences that you have when you are in a state of ignorance? Are preferences satisfied after you die to count? I’ll give you one example: John Rawls talks about somebody who’s ultimate preference is to count the number of blades of grass in a lawn. It’s not that that will make him really happy if he does it, and he’s not ignorant or misinformed – it’s just that that is what he wants to do. Should we really think that that is an important preference to satisfy?'' Derek Parfit raises another challenge to the preference view. If preference maximizers could obtain pills that were cheap and brought no pleasure, yet were highly addictive, then they ought to get addicted to the pills, as more preferences would be satisfied.
        As a concrete example of preferences differing from wellbeing, recall that an individual following their preferences to browse social media all day may not actually be maximizing their short-term or long-term wellbeing \citep{kross2013facebook}. Thus, choices do not perfectly reveal preferences, and preferences do not always correspond to wellbeing, as illustrated in \Cref{fig:wellbeing_compare}. This distinction was popularized by Nobel Prize laureate Amartya Sen \citep{anderson2001symposium}.
    \end{enumerate}
\end{enumerate}

\end{document}